\documentclass{article} 
\usepackage{jvlm_datafm_iclr2026,times}

\usepackage{amsmath,amsfonts,bm}

\def\eqref#1{equation~\ref{#1}}

\def\1{\bm{1}}

\DeclareMathAlphabet{\mathsfit}{\encodingdefault}{\sfdefault}{m}{sl}
\SetMathAlphabet{\mathsfit}{bold}{\encodingdefault}{\sfdefault}{bx}{n}

\usepackage{hyperref}
\usepackage{url}
\usepackage{multirow}
\usepackage{graphicx}
\usepackage{booktabs}
\usepackage{array}
\usepackage{longtable}
\usepackage{float}     
\usepackage{placeins}  
\usepackage{siunitx}   
\usepackage[ruled,vlined]{algorithm2e}

\newcolumntype{L}{>{\raggedright\arraybackslash}p{0.25\textwidth}}
\newcolumntype{R}{>{\raggedright\arraybackslash}p{0.72\textwidth}}
\newcommand{\ds}[1]{\texttt{#1}}
\newcommand{\jvlm}{\href{https://huggingface.co/jinaai/jina-vlm}{\texttt{jina-vlm}}}

\usepackage[most]{tcolorbox}
\tcbset{breakable=false}
\usepackage{caption} 

\newtcolorbox{sftpanel}[1]{%
  enhanced,
  unbreakable,
  colback=gray!2,
  colframe=gray!60,
  boxrule=0.4pt,
  arc=1.5mm,
  outer arc=1.5mm,
  left=1em,
  right=1em,
  top=0.8em,
  bottom=0.8em,
  fonttitle=\large\bfseries,
  title={#1}
}

\title{\jvlm{}: Small Multilingual Vision Language Model}

\author{\textbf{Andreas Koukounas}\thanks{Equal contribution.} \quad
  \textbf{Georgios Mastrapas}\footnotemark[1] \quad
  \textbf{Florian H\"onicke} \quad
  \textbf{Sedigheh Eslami}\thanks{Work done during internship at Jina AI.} \\
  \textbf{Guillaume Roncari} \quad
  \textbf{Han Xiao} \\[1em]
  Jina AI \textit{by} Elastic \\
  Prinzessinnenstr. 19-20, Berlin 10969, Germany\\
  \texttt{research@jina.ai}
}

\iclrfinalcopy 

\begin{document}

\maketitle


\begin{abstract}

We present \jvlm{}, a token-efficient 2.4B parameter vision-language model that achieves state-of-the-art multilingual VQA performance among open 2B-scale VLMs. The model couples a SigLIP2 vision encoder with a Qwen3 language decoder and makes use of image tiling and attention-pooling for token-efficient processing of arbitrary-resolution images. To understand the contribution of different training data categories, we conduct a leave-one-out data mixture ablation study—systematically removing task, domain, modality, and language categories—to diagnose which data types are necessary versus redundant and whether task benefits transfer across domains. Model weights and code are publicly released at \url{https://huggingface.co/jinaai/jina-vlm}.

\end{abstract}


\section{Introduction}
\label{sec:introduction}

Vision-language models (VLMs) combine pretrained vision encoders with large language models to tackle tasks requiring joint visual and textual understanding \citep{flamingo, llava}. Recent VLMs have achieved strong results on visual question answering (VQA), OCR, and multimodal reasoning. However, two challenges limit their practical deployment. First, multilingual capabilities often degrade during vision adaptation: models that perform well on English benchmarks show uneven results across other languages \citep{vlmsurvey}. Second, high-quality VLMs remain computationally expensive to train and deploy, limiting accessibility for researchers and practitioners with constrained resources.

This work introduces \jvlm{}, a 2.4B parameter VLM that addresses both challenges. The model aligns a \texttt{SigLIP2-So400M/14-384} vision encoder \citep{siglip2} with \texttt{Qwen3-1.7B-Base} \citep{qwen3} through an attention-pooling connector, trained with a two-stage pipeline that explicitly incorporates multilingual data. Among open 2B-scale VLMs, \jvlm{} achieves state-of-the-art multilingual performance on MMMB and Multilingual MMBench \citep{mmmb}. On standard English benchmarks spanning diagrams, charts, documents, and OCR, \jvlm{} achieves the highest average score (72.3) across eight VQA benchmarks among 2B-scale VLMs. These results are enabled by a training recipe that incorporates high-quality English and multilingual data. Token efficiency is achieved via an arbitrary-resolution pipeline that combines overlapping tiling with attention-based token pooling to reduce visual token count by 4$\times$.

We identify dataset mixture optimization as a key consideration in training small-scale VLMs. Prior work focuses on total data volume and quality or domain diversity~\citep{cambrian1, mm1, eagle2}, concentrating less on the fine-grained data categories (task type, domain, language) and whether their benefits transfer across tasks. To this end, we conduct an ablation study to provide preliminary insights into which data types are essential versus redundant and whether data benefits transfer across domains and tasks.


\section{Related Work}
\label{sec:related-work}

\textbf{VLM architecture and training}. Modern VLMs follow an architectural pattern in which a pretrained vision encoder extracts visual features, a connector projects them into the language model's embedding space, and a language model generates text conditioned on these visual tokens. This pattern was introduced by PaLI \citep{pali} using an encoder-decoder language model, and popularized in its decoder-only form by LLaVA \citep{llava}. Vision Transformers (ViTs) \citep{vit} produce patch-level representations that the language model processes alongside text embeddings. This decoder-only design is adopted by LLaVA \citep{llava, llava_1_5, llava_uhd, llava_next_interleave, llava_onevision}, QwenVL \citep{qwen_vl, qwen_2_vl, qwen_2_5_vl}, InternVL \citep{internvl, internvl_1_5, internvl_2_5, internvl_3, internvl_3_5}, and Ovis \citep{ovis, ovis_2_5}. Training strategies vary: \citet{qwen_2_vl, internvl_2_5} mix multimodal and text-only data across training stages; \citet{llava_1_5} incorporate academic VQA datasets; \citet{molmo}, \citet{llava_onevision}, and \citet{cambrian1} curate large-scale, diverse data mixtures.

\textbf{Efficient resolution-agnostic image processing}. Standard ViTs process fixed-resolution images, requiring resizing that discards fine-grained detail. Since visual token count scales with resolution and Transformer computation scales quadratically with sequence length, naive high-resolution processing is prohibitive. Several solutions exist: \citet{molmo} tile images with overlap; \citet{qwen_2_vl} introduce Naive Dynamic Resolution with Multimodal Rotary Positional Embedding \citep{rope, 2drope}; \citet{ovis_2_5} use native-resolution ViTs \citep{navit}. Orthogonally, images often contain low-information regions (e.g., sky backgrounds), making visual tokens highly redundant. Token compression methods address this \citep{fastv,prumerge,visionzip,pyramiddrop}. \citet{internvl_1_5} develop Dynamic High-Resolution Tiling, and \citet{nvila} propose scale-then-compress strategies. Recent work on training-free token budgeting, such as HERO \citep{hero}, demonstrates that inference-time pruning can achieve significant speedups while preserving accuracy; our approach differs by learning compact representations during training rather than dropping tokens at inference.

\textbf{Vision-language connectors}. The connector bridging vision encoders and language models significantly impacts both efficiency and performance. BLIP-2 \citep{blip2} introduces Q-Former, a learnable query-based transformer that extracts fixed-length representations from visual features, reducing the number of tokens fed to the LLM. Flamingo \citep{flamingo} uses a Perceiver Resampler with cross-attention to compress visual tokens. Our attention-pooling connector reduces tokens but operates differently: rather than learning a fixed set of queries, we apply local 2$\times$2 attention pooling that preserves spatial structure while achieving 4$\times$ compression. Dense Connector~\citep{denseconnector} concatenates features from multiple ViT layers and shows that intermediate layers carry complementary spatial and semantic information that improves VLM performance. MM1~\citep{mm1} studies the effect of vision encoder configuration, including image resolution, token count, and connector design, on downstream accuracy. Our connector design draws on these observations, concatenating an intermediate and a near-final layer to capture both fine-grained and high-level features, combined with attention pooling for token reduction.

\textbf{Small VLMs}. Efficiency has become a central objective. \citet{imp} combine quantization with aggressive resolution reduction for mobile deployment, matching larger models' performance. MiniCPM-V \citep{minicpmv} targets edge deployment while maintaining strong OCR and multilingual capabilities. \citet{smolvlm} systematically explore design parameters to train VLMs as small as 256M parameters.

\textbf{Multilinguality and text-only performance}. Many lightweight VLMs \citep{paligemma,paligemma2,phi3} achieve strong English performance but degrade on other languages or on text-only capabilities. \citet{qwen_2_vl} and \citet{internvl_1_5} improve on non-English languages through targeted multilingual training data, and \citet{pangea} introduce instruction-tuning data spanning 39 languages. For pure text performance, mitigation strategies include parameter-efficient fine-tuning \citep{cauldron} and staged training with progressive unfreezing \citep{llava_onevision}.

\textbf{Data mixture analysis}. Several recent works investigate the role of training data composition in VLM performance. MM1~\citep{mm1} conducts ablations over image-caption, interleaved, and text-only data ratios, finding that mixing all three modalities and carefully tuning their proportions is critical for strong few-shot performance; they also study the effect of image resolution and token count on downstream accuracy. Cambrian-1~\citep{cambrian1} introduces a large-scale data curation effort and evaluates the effect of data source diversity on visual grounding, reporting that balanced sampling across sources outperforms naive concatenation. Molmo/PixMo~\citep{molmo} demonstrate that high-quality, targeted caption data can substitute for much larger but noisier mixtures, and provide insights on how caption style and granularity affect downstream task transfer. Eagle~2~\citep{eagle2} proposes a post-training data strategy built on diversity-first collection and quality filtering, using K-means clustering for balanced representation and rule-based removal of low-quality samples, showing gains from careful data curation rather than data scaling. Our work is complementary: rather than optimizing data proportions or sample weights, we perform \emph{leave-one-out} ablations that remove entire data categories to measure their marginal contribution and cross-domain transfer effects. This provides a different diagnostic lens, identifying which categories are necessary versus redundant, that is particularly relevant for practitioners operating under fixed compute budgets.


\section{Model Architecture}
\label{sec:model-architecture}

\begin{figure}[t!]
    \centering
    \includegraphics[width=\linewidth]{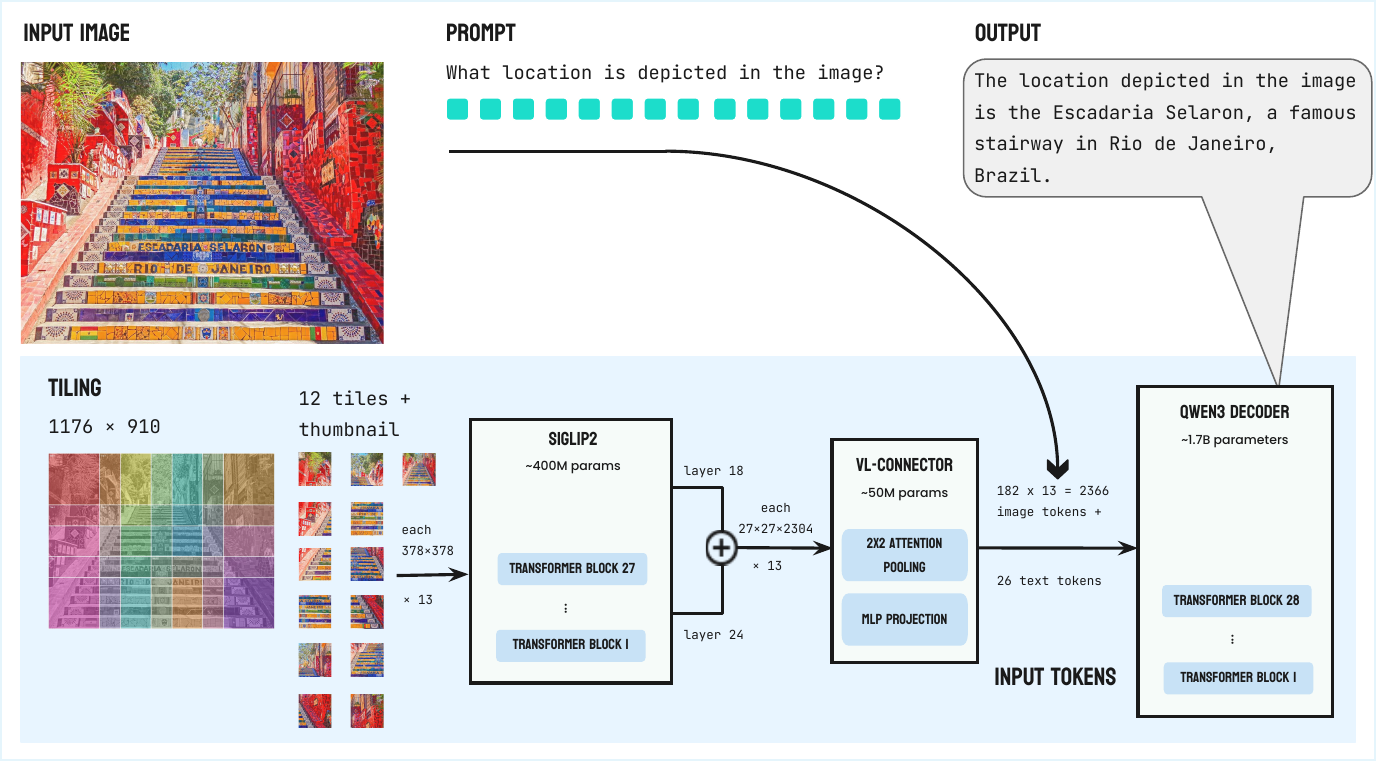}
    \caption{Architecture of \jvlm. Images are resized to fit a grid of up to 12 overlapping tiles, plus a global thumbnail. Each tile is a square 378$\times$378 crop; adjacent tiles overlap by 112 pixels with a stride of 266 pixels between tile origins. A 4$\times$3 grid therefore spans 1176$\times$910 pixels, and images exceeding this effective resolution are downscaled to fit the tile budget. Each tile produces 729 patches via SigLIP2 \citep{siglip2}. The VL connector concatenates features from layers 24 and 18, the third- and ninth-to-last layers, then applies 2$\times$2 attention pooling to reduce 729 tokens to 182 before projecting to the decoder dimension. Visual tokens are combined with text embeddings for the Qwen3 decoder \citep{qwen3}.}
    \label{fig:model}
\end{figure}

Figure \ref{fig:model} illustrates the architecture of \jvlm{}. The model uses overlapping image tiling following \citet{molmo}, combined with attention-based token pooling to reduce sequence length while preserving spatial information.

\subsection{Vision Encoder}
\label{sec:vision-encoder}

The vision encoder, \texttt{SigLIP2-So400M/14-384}\footnote{\url{https://huggingface.co/google/siglip2-so400m-patch14-384}}, is a 27-layer Vision Transformer with 400M parameters that processes 378$\times$378 pixel inputs as 27$\times$27 grids of 14$\times$14 patches. To handle arbitrary resolutions, we decompose each image into overlapping tiles of this size and process each tile independently through the encoder. A global thumbnail, the full image resized to 378$\times$378, provides context alongside the tile representations. We use a default of 12 tiles during training; this limit can be increased at inference or during continued training to handle higher resolutions, with memory scaling linearly with tile count. The tiling algorithm is detailed in Appendix~\ref{appendix:cropping-pseudocode}.

\subsection{Vision-Language Connector}
\label{sec:vision-language-connector}

Rather than using the final ViT output, \jvlm{} concatenates features from two intermediate layers. This design is motivated by prior findings that intermediate ViT layers retain spatial detail that is progressively abstracted in later layers~\citep{denseconnector}. Following \citet{molmo}, we select the third-to-last and ninth-to-last layers (24th and 18th layer), as they span the transition from mid-level spatial to high-level semantic representations in this 27-layer encoder. The connector then applies attention pooling over 2$\times$2 patch neighborhoods, using mean-pooled features as queries, a design inspired by Q-Former~\citep{blip2} and Perceiver Resampler~\citep{flamingo}. This reduces the token count by 4$\times$ while preserving local structure. A SwiGLU projection layer maps the pooled representations to the language model's embedding dimension.

In more formal terms, let $\mathbf{H}^{(\ell)} \in \mathbb{R}^{N \times d_v}$ denote the hidden states from ViT layer $\ell$, where $N$ is the number of patches, $d_v$ is the vision encoder hidden size, and negative indices count from the final layer (e.g., $\ell = -1$ is the last layer). We concatenate features from two internal layers:
\begin{equation}
\mathbf{H}_{\text{concat}} = [\mathbf{H}^{(-3)}; \mathbf{H}^{(-9)}] \in \mathbb{R}^{N \times 2d_v}
\end{equation}

For each $2{\times}2$ patch neighborhood $\mathcal{N}_i$, we compute a query vector as the mean of the neighborhood features:
\begin{equation}
\mathbf{q}_i = \frac{1}{4}\sum_{j \in \mathcal{N}_i} \mathbf{h}_j, \quad
\mathbf{Q} = [\mathbf{q}_1; \dots; \mathbf{q}_M] \in \mathbb{R}^{M \times 2d_v}
\end{equation}
where $\mathcal{N}_i$ contains the four patches at positions $(2i_x, 2i_y)$, $(2i_x+1, 2i_y)$, $(2i_x, 2i_y+1)$, and $(2i_x+1, 2i_y+1)$, and $M = N/4$.

Attention pooling is then computed as:
\begin{equation}
\mathbf{H}_{\text{pooled}} = (\text{softmax}\left(\frac{\mathbf{Q}\mathbf{W}_Q(\mathbf{H}_{\text{concat}}\mathbf{W}_K)^\top}{\sqrt{d_k}}\right)\mathbf{H}_{\text{concat}}\mathbf{W}_V) \mathbf{W}_O \in \mathbb{R}^{M \times d_v}
\end{equation}
where $d_k=d_v$ and $\mathbf{W}_Q \in \mathbb{R}^{2d_v \times d_k}$, $\mathbf{W}_K \in \mathbb{R}^{2d_v \times d_k}$, $\mathbf{W}_V \in \mathbb{R}^{2d_v \times 2d_v}$ and $\mathbf{W}_O \in \mathbb{R}^{2d_v \times d_v}$ are learnable weight matrices. Finally, the pooled visual features are projected to the language model embedding dimension via a SwiGLU \citep{swiglu} layer:
\begin{equation}
\mathbf{H}_{\text{proj}} = \left( \text{Swish}(\mathbf{H}_{\text{pooled}}\mathbf{W}_1) \odot (\mathbf{H}_{\text{pooled}}\mathbf{W}_2) \right) \mathbf{W}_3 \in \mathbb{R}^{M \times d_l}
\end{equation}
where $\text{Swish}(x) = x \cdot \sigma(x)$, $\sigma$ is the sigmoid function, $\odot$ denotes element-wise multiplication, $\mathbf{W}_1, \mathbf{W}_2 \in \mathbb{R}^{d_v \times 3d_l}$, $\mathbf{W}_3 \in \mathbb{R}^{3d_l \times d_l}$ are learnable parameters, and $d_l$ is the language model embedding size.

\subsection{Language Decoder}
\label{sec:language-decoder}

The language decoder is initialized from \texttt{Qwen3-1.7B-Base}\footnote{\url{https://huggingface.co/Qwen/Qwen3-1.7B-Base}}, which empirically outperformed the instruction-tuned variant in our setting. We introduce three special tokens to structure visual inputs: \texttt{<im\_start>} and \texttt{<im\_end>} delimit image and thumbnail sequences, while \texttt{<im\_col>} marks row boundaries within the patch grid, where tokens are arranged left-to-right and top-to-bottom. Input and output embedding weights are not tied.

\subsection{Efficiency Analysis}
\label{sec:efficiency-analysis}

Table~\ref{tab:efficiency} quantifies the computational benefits of attention pooling. With the default 12-tile configuration (plus thumbnail), the unpooled baseline would produce 9,477 visual tokens per image, while our 2$\times$2 pooling reduces this to 2,366 tokens. Since the ViT processes each tile identically regardless of pooling, the savings apply exclusively to the LLM: we observe a 3.9$\times$ reduction in prefill FLOPs and a 4$\times$ reduction in KV-cache memory. The overall FLOPs reduction is 2.3$\times$ when including the shared ViT cost.

\begin{table}[htpb]
\centering
\caption{Efficiency comparison with and without 2$\times$2 attention pooling for the default 12-tile configuration. FLOPs are computed for LLM prefill; KV-cache assumes fp16 precision.}
\begin{tabular}{lccc}
\toprule
Metric & No Pooling & With Pooling & Reduction \\
\midrule
Visual tokens & 9,477 & 2,366 & 4.0$\times$ \\
LLM prefill FLOPs & 27.2 TFLOPs & 6.9 TFLOPs & 3.9$\times$ \\
KV-cache memory & 2.12 GB & 0.53 GB & 4.0$\times$ \\
\bottomrule
\end{tabular}
\label{tab:efficiency}
\end{table}


\section{Training}
\label{sec:training}

Training proceeds in two stages, with both stages updating all model components (encoder, connector, and decoder) without freezing, following \citet{molmo}. The combined data comprises samples in multiple languages including English, Chinese, Arabic, German, Spanish, French, Italian, Japanese, Korean, Portuguese, Russian, Turkish, Vietnamese, Thai, Indonesian, Hindi, and Bengali. Table~\ref{tab:hyperparams} summarizes hyperparameters for both stages.

\subsection{Stage 1: Alignment Training}
\label{sec:alignment-training}

The first stage focuses on cross-language semantic grounding rather than task-specific objectives. Training data consists primarily of caption datasets (\texttt{PixmoCap} \citep{molmo}, \texttt{PangeaIns} \citep{pangea}) spanning diverse visual domains: natural scenes, documents, infographics, and diagrams. Following \citet{mm1}, who show that mixing text-only data during multimodal training preserves language capabilities, we include 15\% text-only data from \texttt{PleiAS/common\_corpus} \citep{pleias} to mitigate degradation on text-only tasks. The connector uses a higher learning rate and shorter warmup than the encoder and decoder.

\subsection{Stage 2: Instruction Fine-tuning}
\label{sec:instruction-finetuning}

The second stage trains instruction-following for VQA and reasoning tasks. We combine public dataset collections, including \texttt{LLaVA OneVision} \citep{llava_onevision}, \texttt{Cauldron} \citep{cauldron}, \texttt{Cambrian} \citep{cambrian1}, \texttt{PangeaIns} \citep{pangea}, and \texttt{FineVision} \citep{finevision}, with text-only instruction data from \citet{aya}. The mixture covers academic VQA, document understanding, OCR, mathematics, and reasoning. Appendix~\ref{appendix:examples} shows representative examples.

Given the diversity of instruction data, we found single-source batches more effective initially, likely due to the heterogeneous data mixture. We train for 30K steps with single-source batches, then 30K steps with mixed-source batches.

\begin{table}[!t]
\centering
\caption{Model training hyperparameters across pre-training and fine-tuning stages.}
\begin{tabular}{lcc}
\toprule
Hyperparameter & Pre-Training & Fine-Tuning \\
\midrule
Warmup ViT & 10\% & 10\% \\
Warmup Con. & 1\% & 10\% \\
Warmup LLM & 10\% & 10\% \\
LR ViT & 6e-6 & 5e-6 \\
LR Con. & 2e-4 & 5e-6 \\
LR LLM & 2e-5 & 1e-5 \\
Cosine Decay & 0.1 & 0.1 \\
Eps. & 1e-6 & 1e-6 \\
Betas & 0.9, 0.95 & 0.9, 0.95 \\
Batch Size & 128 & 256 \\
Steps & 25K & 60K \\
Samples & 3.2M & 15.3M \\
Tokens & 10B & 37B \\
GPU Hours & 296 & 1,000 \\
\bottomrule
\end{tabular}
\label{tab:hyperparams}
\end{table}


\section{Evaluation}
\label{sec:evaluation}

We compare \jvlm{} against lightweight VLMs across six capability areas: general VQA, multimodal comprehension, multi-image reasoning and hallucination control, mathematical reasoning, text-only performance, and multilingual understanding. All evaluations use VLMEvalKit\footnote{\href{https://github.com/open-compass/VLMEvalKit}{\texttt{https://github.com/open-compass/VLMEvalKit}}} \citep{vlmevalkit} with English prompts matching our training format (e.g., ``Return only the letter of the best answer option'' for multiple-choice, ``Respond very briefly'' for open-ended questions).

\subsection{General VQA Tasks}
\label{sec:evaluation-vqa}

\begin{table}[!t]
\centering
\caption{Comparison of general visual question answering performance.}
\resizebox{\textwidth}{!}{%
\begin{tabular}{l S[table-format=2.1] S[table-format=2.1] S[table-format=2.1] S[table-format=2.1,table-space-text-post={*}] S[table-format=2.1,table-space-text-post={*}] S[table-format=3.0] S[table-format=2.1,table-space-text-post={*}] c S[table-format=2.1]}
\toprule
{Model} & {AI2D} & {ChartQA} & {TextVQA} & {DocVQA} & {InfoVQA} & {OCR} & {SEED-2} & {CharXiv} & {Overall} \\
 & & {(test avg)} & {(val)} & {(val)} & {(val)} & {Bench} & {Plus} & {(RQ / DQ)} & \\
\midrule
{\jvlm{}}       & \bfseries 82.0 & \bfseries 81.9 & \bfseries 83.2 & 90.6 & 71.6 & 778 & 67.2 & {\bfseries 32.3} / 63.5 & \bfseries 72.3 \\
{Qwen2-VL-2B}   & 74.7 & 73.5 & 79.7 & 89.2{*} & 64.0{*} & 809 & 62.4 & 23.3 / 55.0{*} & 66.4 \\
{Qwen3-VL-2B}   & 76.9 & 77.2 & 79.5 & \bfseries 92.3{*} & \bfseries 71.9{*} & \bfseries 858 & 67.3{*} & 28.8 / 62.3 & 71.6 \\
{InternVL3-2B}  & 78.6 & 80.2 & 77.0 & 87.4{*} & 67.1{*} & 835 & 64.6 & 28.3 / 54.7 & 69.2 \\
{InternVL3.5-2B} & 78.8 & 80.7 & 76.5 & 88.5{*} & 69.3{*} & 836 & \bfseries 68.0 & 31.6 / {\bfseries 65.0} & 71.6 \\
\bottomrule
\end{tabular}
}
\raggedright\small
~~~
\newline
Results for models other than \jvlm{} are from their respective papers \citep{internvl_3_5, internvl_3, qwen_2_vl}, except those marked with * which were computed using VLMEvalKit. All scores represent accuracy (\%) except OCRBench which uses a 0--1000 scale; for overall average computation, OCRBench scores are divided by 10 to align with the 0--100 scale of other benchmarks.
\label{tab:model_vl_benchmark}
\end{table}

Table~\ref{tab:model_vl_benchmark} reports results on eight VQA benchmarks covering diagrams (AI2D \citep{ai2d}), charts (ChartQA \citep{chartqa}, CharXiv \citep{charxiv}), scene text (TextVQA \citep{textvqa}), documents (DocVQA \citep{docvqa}, InfoVQA \citep{infovqa}), OCR (OCRBench \citep{ocrbench}), and diverse scenes (SEED-Bench-2-Plus \citep{seedbench2plus}). \jvlm{} achieves the highest average (72.3), with particularly strong performance on diagram interpretation and text extraction.

\subsection{Multimodal and Real-World Understanding}
\label{sec:evaluation-multimodal-realworld}

\begin{table}[!t]
\centering
\caption{Comparison of generic multimodal understanding and real-world understanding performance.}
\resizebox{\textwidth}{!}{%
\begin{tabular}{l S[table-format=4.1,table-space-text-post={*}] S[table-format=2.1] S[table-format=2.1] S[table-format=2.1] S[table-format=2.1] S[table-format=2.1,table-space-text-post={*}] S[table-format=2.1,table-space-text-post={*}] S[table-format=2.1,table-space-text-post={*}]}
\toprule
{Model} & {MME} & {MMB v1.1} & {MMStar} & {Overall} & {RealWorld} & {MME-RW} & {R-Bench} & {Overall} \\
 & {(sum)} & {(EN)} & & {(MM)} & {QA} & {(EN)} & {(dis)} & {(RW)} \\
\midrule
{\jvlm{}}      & 1965.8 & 75.8 & 56.2 & 67.4 & \bfseries 68.2 & 50.7 & 66.7 & 61.9 \\
{Qwen2-VL-2B}  & 1872.0 & 72.2 & 48.0 & 62.4 & 62.9 & 38.7{*} & 63.2 & 55.0{*} \\
{Qwen3-VL-2B}  & 2000.8{*} & 77.8 & 58.3 & 69.2 & 63.9 & \bfseries 57.9{*} & 67.3{*} & \bfseries 63.0 \\
{InternVL3-2B} & \bfseries 2221.2 & \bfseries 78.6 & 60.7 & \bfseries 72.9 & 64.3 & 53.8 & \bfseries 67.5 & 61.9 \\
{InternVL3.5-2B} & 2123.3 & 76.6 & \bfseries 62.7 & 71.7 & 62.0 & 49.7 & 62.4 & 58.0 \\
\bottomrule
\end{tabular}
}
\raggedright\small
~~~
\newline
Results for models other than \jvlm{} are from their respective papers \citep{internvl_3_5, internvl_3, qwen_2_vl}, except those marked with * which are computed using VLMEvalKit. All scores represent accuracy (\%) except MME which uses a 0--2800 scale; for overall average computation, MME scores are divided by 28 to align with the 0--100 scale of other benchmarks.
\label{tab:multimodal-realworld-evals}
\end{table}

Table~\ref{tab:multimodal-realworld-evals} shows results on multimodal comprehension (MME \citep{mme}, MMB v1.1 \citep{mmbench}, MMStar \citep{mmstar}) and real-world understanding (RealWorldQA \citep{realworldqa}, MME-RealWorld \citep{mmerealworld}, R-Bench \citep{rbench}). \jvlm{} scores 67.4 on multimodal tasks and 61.9 on real-world tasks, achieving the best RealWorldQA result (68.2).

\subsection{Multi-Image Reasoning and Hallucination}
\label{sec:evaluation-multiimage-hallucination}

\begin{table}[ht]
\centering
\caption{Comparison of multi-image and hallucination performance.}
\begin{tabular}{l S[table-format=2.1] S[table-format=2.1,table-space-text-post={*}] S[table-format=2.1,table-space-text-post={*}] S[table-format=2.1] S[table-format=2.1] S[table-format=2.1,table-space-text-post={*}] S[table-format=2.1]}
\toprule
{Model} & {BLINK} & {Muir} & {MMT} & {Overall} & {HallBench} & {POPE} & {Overall} \\
 & {(val)} & {Bench} & {(val)} & {(MI)} & {(avg)} & {(avg)} & {(Hall)} \\
\midrule
{\jvlm{}}      & 50.1 & 34.7 & 57.2 & 47.3 & 39.1 & \bfseries 90.3 & 64.7 \\
{Qwen2-VL-2B}  & 44.4 & 25.5{*} & 55.1 & 41.7 & 41.7 & 87.9{*} & 64.8 \\
{Qwen3-VL-2B}  & \bfseries 53.8 & \bfseries 47.4 & \bfseries 60.0{*} & \bfseries 53.7 & 44.5 & 88.9{*} & 66.7 \\
{InternVL3-2B} & 50.3 & 38.8 & 59.5 & 49.5 & 42.5 & 89.6 & 66.1 \\
{InternVL3.5-2B} & 51.3 & 44.0 & 58.5 & 51.3 & \bfseries 48.6 & 87.2 & \bfseries 67.9 \\
\bottomrule
\end{tabular}
\raggedright\small
~~~
\newline
Results for models other than \jvlm{} are from their respective papers, \citep{internvl_3_5, internvl_3, qwen_2_vl}, except those marked with * which are computed using VLMEvalKit. All scores represent accuracy (\%).
\label{tab:multiimage-hallucination-evals}
\end{table}

Table~\ref{tab:multiimage-hallucination-evals} reports multi-image reasoning (BLINK \citep{blink}, MuirBench \citep{muirbench}, MMT \citep{mmtbench}) and hallucination benchmarks that measure the tendency to fabricate visual details (HallBench \citep{hallusionbench}, POPE \citep{pope}). \jvlm{} scores 47.3 on multi-image tasks, which is expected given limited multi-image training data, but achieves the best POPE score (90.3), indicating low hallucination rates.

\subsection{Mathematical Reasoning}
\label{sec:evaluation-math-reasoning}

\begin{table}[!t]
\centering
\caption{Comparison of multimodal reasoning and mathematical problem-solving performance.}
\resizebox{\textwidth}{!}{%
\begin{tabular}{l S[table-format=2.1] c c c c c S[table-format=2.1]}
\toprule
{Model} & {MMMU} & {MathVista} & {MathVision} & {MathVerse} & {WeMath} & {LogicVista} & {Overall} \\
 & & & & {(Vision Only)} & & & \\
\midrule
{\jvlm{}}      & 45.6 & 59.5 & 19.2 & 23.9 & 17.1 & 33.3 & 33.1 \\
{Qwen2-VL-2B}  & 41.1 & 43.0 & 12.4 & 17.3{*} & 10.9{*} & 27.3{*} & 25.3 \\
{Qwen3-VL-2B}  & 53.4 & 61.3 & 31.6 & 22.7{*} & 28.0{*} & 35.4{*} & 38.7 \\
{InternVL3-2B} & 48.6 & 57.0 & 21.7 & 25.3 & 22.4 & 36.9 & 35.3 \\
{InternVL3.5-2B} & \bfseries 59.0 & {\bfseries 71.8} / 61.5$^\dagger$ & {\bfseries 42.8} / 26.5$^\dagger$ & {\bfseries 53.4} / 35.3$^\dagger$ & {\bfseries 48.5} / 19.1$^\dagger$ & {\bfseries 47.7} / 41.4$^\dagger$ & \bfseries 50.7 \\
\bottomrule
\end{tabular}
}
\raggedright\small
~~~
\newline
Results for models other than \jvlm{} are from their respective papers, \citep{internvl_3_5, internvl_3, qwen_2_vl}, except those marked with * which are computed using VLMEvalKit. $\dagger$ indicates scores for \texttt{InternVL3.5-2B} without thinking mode, evaluated using VLMEvalKit. All scores represent accuracy (\%).
\label{tab:model_math_benchmark}
\end{table}

Table~\ref{tab:model_math_benchmark} reports structured reasoning benchmarks: multidisciplinary comprehension (MMMU \citep{mmmu}), visual mathematics (MathVista \citep{mathvista}, MathVision \citep{mathvision}, MathVerse \citep{mathverse}, WeMath \citep{wemath}), and logical reasoning (LogicVista \citep{logicvista}). \jvlm{} performs comparably to InternVL3-2B and outperforms Qwen2-VL-2B.

\subsection{Text-Only Performance}
\label{sec:evaluation-text-only}

\begin{table}[!t]
\centering
\caption{Comparison of text-only benchmarks.}
\begin{tabular}{l S[table-format=2.1] S[table-format=2.1] S[table-format=2.1] S[table-format=2.1] S[table-format=2.1] S[table-format=2.1]}
\toprule
{Model} & {MMLU} & {MMLU-Pro} & {GSM-8k} & {ARC-C} & {HellaSwag} & {Overall} \\
\midrule
{\jvlm{}}  & 56.1 & 30.3 & 71.3 & \bfseries 77.3 & \bfseries 59.4 & 58.9 \\
{Qwen3-1.7B}  & \bfseries 62.6 & \bfseries 46.4 & \bfseries 75.3 & 73.4 & 59.0 & \bfseries 63.3 \\
\bottomrule
\end{tabular}
\raggedright\small
~~~
\newline
Results are collected using our evaluation code. All scores represent accuracy (\%).
\label{tab:text-only-evals}
\end{table}

Table~\ref{tab:text-only-evals} compares \jvlm{} against the backbone Qwen3-1.7B on text-only benchmarks: MMLU \citep{mmlu}, MMLU-Pro \citep{mmlupro}, GSM-8K \citep{gsm8k}, ARC-C \citep{arc}, and HellaSwag \citep{hellaswag}. Results show mixed preservation of text-only capabilities: \jvlm{} matches or exceeds the backbone on commonsense reasoning (ARC-C, HellaSwag) and retains most performance on MMLU and GSM-8K. However, MMLU-Pro shows substantial degradation (46.4 $\rightarrow$ 30.3), likely because this benchmark emphasizes extended multi-step reasoning that conflicts with our instruction-tuning toward concise visual responses.

\subsection{Multilingual Understanding}
\label{sec:evaluation-multilingual}

\begin{table}[t!]
\centering
\caption{Comparison of multilingual multimodal understanding performance.}
\resizebox{\textwidth}{!}{%
\begin{tabular}{ll S[table-format=2.1] S[table-format=2.1] S[table-format=2.1,table-space-text-post={*}] S[table-format=2.1] S[table-format=2.1]}
\toprule
\multicolumn{2}{c}{Benchmark} & {\jvlm{}} & {Qwen2-VL-2B} & {Qwen3-VL-2B} & {InternVL3-2B} & {InternVL3.5-2B} \\
\midrule
\multirow{7}{*}{MMMB}
 & ar  & \bfseries 76.9 & 68.3 & 72.7{*} & 68.6 & 68.5 \\
 & cn  & \bfseries 80.0 & 74.2 & 75.7{*} & 78.3 & 77.7 \\
 & en  & \bfseries 82.0 & 78.3 & 80.7{*} & 81.9 & 80.2 \\
 & pt  & \bfseries 79.2 & 72.6 & 75.0{*} & 75.4 & 75.9 \\
 & ru  & \bfseries 79.2 & 72.8 & 75.9{*} & 74.6 & 76.3 \\
 & tr  & \bfseries 75.5 & 61.8 & 68.5{*} & 62.9 & 69.1 \\
 & \textit{avg} & \bfseries 78.8 & 71.3 & 75.0{*} & 73.6 & 74.6 \\
\midrule
\multirow{7}{*}{\shortstack{Multi.\\MMBench}}
 & ar  & \bfseries 70.0 & 66.7 & 66.2{*} & 66.4 & 63.7 \\
 & cn  & 75.9 & 67.0 & 75.7{*} & \bfseries 77.8 & 75.9 \\
 & en  & 78.8 & 71.1 & 77.8{*} & \bfseries 81.3 & 78.4 \\
 & pt  & 74.7 & 72.1 & 71.4{*} & \bfseries 75.9 & 73.7 \\
 & ru  & 75.3 & 69.9 & \bfseries 75.9{*} & 70.7 & 71.4 \\
 & tr  & \bfseries 71.1 & 69.3 & 67.0{*} & 59.5 & 62.0 \\
 & \textit{avg} & \bfseries 74.3 & 69.4 & 72.3{*} & 71.9 & 70.9 \\
\midrule
\multicolumn{2}{l}{MTVQA} & 25.6 & 20.6 & 27.3{*} & 26.7 & \bfseries 28.5 \\
\midrule
\multicolumn{2}{l}{Overall} & \bfseries 59.6 & 53.8 & 58.2 & 57.4 & 58.0 \\
\bottomrule
\end{tabular}
}
\raggedright\small
~~~
\newline
Results for baseline models are derived from their original publications~\citep{internvl_3_5, internvl_3, qwen_2_vl}, except those marked with * which are computed using VLMEvalKit. All scores represent accuracy (\%).
\label{tab:multilingual-multimodal-evaluation}
\end{table}

Table~\ref{tab:multilingual-multimodal-evaluation} reports multilingual multimodal benchmarks: MMMB \citep{mmmb}, Multilingual MMBench \citep{mmmb}, and MTVQA \citep{mtvqa}. \jvlm{} achieves state-of-the-art multilingual performance among 2B-scale VLMs, with the highest averages on MMMB (78.8) and Multilingual MMBench (74.3). In our evaluation setup, the system prompt is in English (following the training format, not a specific deployment target) while the questions, answer options, and text within images are presented in the target language. Baseline models use their default prompts. This setup tests cross-lingual visual understanding but not target-language instruction-following.


\section{Data Mixture Analysis}
\label{sec:data-mixture-analysis}

The rapid advancement of large-scale VLMs has been driven by scaling both model size and training data. Aggregate data volume and domain diversity have received significant attention~\citep{cambrian1, molmo, eagle2}; however, the composition and relative importance of different data types within training mixtures remain poorly understood. This issue is particularly pressing when training small-scale models, where limited budget needs to be spent optimally. In such settings, we hypothesize that, besides data quality, dataset mixture composition is critical.

We conduct a comprehensive ablation study that systematically removes different data types from our training mixture and measures the downstream impact across diverse capabilities. By training 9 variants of our model, each excluding a specific category of data—such as OCR datasets, VQA data, or multilingual content—we can quantify both the necessity of each component and potential cross-domain transfer effects. Our approach provides diagnostic insights into data category necessity and cross-domain transfer, and offers preliminary empirical evidence for mixture optimization under resource constraints.

\subsection{Experimental Design}
\label{sec:experimental-design}

Our training dataset consists of data across two stages, alignment and instruction fine-tuning, as described in sections~\ref{sec:alignment-training} and~\ref{sec:instruction-finetuning}. Besides dividing by stage, we group data along four axes: modality (multimodal vs.\ text-only), language (English vs.\ multilingual), task type (VQA, OCR, open-ended, reasoning, etc.), and domain (real-world, math, charts, documents, code, etc.). Language detection is done automatically using a language classifier, and modalities are derived from each dataset's schema. Domain and task type categorization is done through manual inspection of examples in each dataset, after defining a high-level taxonomy. Tables~\ref{tab:composition-finetuning},~\ref{tab:task-reference} and~\ref{tab:domain-reference} in Appendix~\ref{appendix:composition-finetuning} show the detailed data mixture composition for the instruction fine-tuning stage.

Starting from the original training mixture, we design 9 experiments, each removing one category---\textbf{Task:} \texttt{Xvqa} (VQA data), \texttt{Xocr} (OCR data); \textbf{Domain:} \texttt{Xrealworld} (real-world photos), \texttt{Xmath} (math/reasoning), \texttt{Xcharts} (charts/diagrams), \texttt{Xdocs} (documents), \texttt{Xcode} (code); \textbf{Modality \& Language:} \texttt{Xtext} (text-only data), \texttt{Xmlingual} (non-English data). All ablations run against the baseline \texttt{base} that uses the original data mixture used to train \jvlm.

To enable efficient exploration within computational constraints, all experiments, including the baseline, run for 10\% of the full training duration. This is a known limitation: ablations at 10\% may not fully reflect final-training behavior, and findings should be treated as diagnostic observations rather than definitive conclusions. Ablation runs do not adjust steps to match the compute of the baseline; when a data category is removed, remaining data may be seen more frequently within the fixed budget. Training order is held fixed across all ablations; disentangling order from composition effects would require additional controlled experiments. Hyperparameters are fixed across all runs and the same two-stage training recipe is used. Each ablation is applied consistently across both stages when applicable (for example, \texttt{Xmlingual} removes non-English data from both alignment and instruction stages). Training checkpoints are evaluated against the same comprehensive evaluation suite described in Section~\ref{sec:evaluation}.

\subsection{Results}
\label{sec:ablation-results}

\begin{table}[t!]
\centering
\caption{Ablation study results.}
\resizebox{\textwidth}{!}{%
\begin{tabular}{l|ccccc}
\toprule
\textbf{Model} & \textbf{AI2D} & \textbf{ChartQA} & \textbf{DocVQA} & \textbf{InfoVQA} & \textbf{OCRBench} \\
\midrule
\texttt{base} & 72.0 {(+0.0)} & 64.5 {(+0.0)} & 81.8 {(+0.0)} & 63.2 {(+0.0)} & 699.0 {(+0.0)} \\
\texttt{Xvqa} & 65.1 {(-9.5)} & 50.2 {(-22.1)} & 75.3 {(-8.0)} & 48.8 {(-22.8)} & 686.0 {(-1.9)} \\
\texttt{Xmlingual} & 74.4 {(+3.3)} & 70.9 {(+9.9)} & 84.0 {(+2.6)} & 64.9 {(+2.6)} & \textbf{736.0 {(+5.3)}} \\
\texttt{Xrealworld} & 74.0 {(+2.8)} & 64.9 {(+0.6)} & \textbf{84.3 {(+3.1)}} & \textbf{65.0 {(+2.8)}} & 717.0 {(+2.6)} \\
\texttt{Xtext} & 72.7 {(+1.1)} & \textbf{70.4 {(+9.1)}} & 84.0 {(+2.7)} & 64.3 {(+1.7)} & 718.0 {(+2.7)} \\
\texttt{Xocr} & \textbf{74.7 {(+3.8)}} & 69.7 {(+8.0)} & 82.3 {(+0.6)} & 64.6 {(+2.2)} & 622.0 {(-11.0)} \\
\texttt{Xcharts} & 65.7 {(-8.7)} & 39.1 {(-39.4)} & 83.0 {(+1.5)} & 63.8 {(+0.8)} & 691.0 {(-1.1)} \\
\texttt{Xdocs} & 74.4 {(+3.5)} & 69.8 {(+8.2)} & 80.6 {(-1.4)} & 63.9 {(+1.1)} & 686.0 {(-1.9)} \\
\texttt{Xmath} & 72.9 {(+1.4)} & 64.9 {(+0.6)} & 83.3 {(+1.8)} & 64.0 {(+1.2)} & 687.0 {(-1.7)} \\
\texttt{Xcode} & 71.6 {(-0.5)} & 64.1 {(-0.7)} & 82.6 {(+1.0)} & 64.0 {(+1.2)} & 700.0 {(+0.1)} \\
\midrule
\textbf{Model} & \textbf{TextVQA} & \textbf{MMStar} & \textbf{MMBench1.1} & \textbf{RealWorldQA} & \textbf{MMMU} \\
\midrule
\texttt{base} & 71.0 {(+0.0)} & 46.7 {(+0.0)} & 69.1 {(+0.0)} & 60.0 {(+0.0)} & 41.0 {(+0.0)} \\
\texttt{Xvqa} & 57.8 {(-18.6)} & 44.8 {(-4.0)} & 66.9 {(-3.2)} & 60.7 {(+1.1)} & 40.6 {(-1.1)} \\
\texttt{Xmlingual} & \textbf{75.5 {(+6.4)}} & 48.3 {(+3.4)} & 70.5 {(+2.1)} & 61.7 {(+2.8)} & 40.9 {(-0.3)} \\
\texttt{Xrealworld} & 72.2 {(+1.7)} & 47.7 {(+2.1)} & 68.8 {(-0.5)} & 59.9 {(-0.2)} & 42.3 {(+3.3)} \\
\texttt{Xtext} & \textbf{75.5 {(+6.3)}} & \textbf{48.6 {(+4.1)}} & 69.8 {(+1.1)} & \textbf{63.8 {(+6.3)}} & 42.6 {(+3.8)} \\
\texttt{Xocr} & 73.5 {(+3.5)} & 48.3 {(+3.4)} & \textbf{71.8 {(+4.0)}} & 60.9 {(+1.5)} & 42.4 {(+3.5)} \\
\texttt{Xcharts} & 72.0 {(+1.5)} & 43.8 {(-6.1)} & 67.5 {(-2.3)} & 62.1 {(+3.5)} & 41.2 {(+0.5)} \\
\texttt{Xdocs} & 75.0 {(+5.6)} & 47.0 {(+0.7)} & 70.9 {(+2.6)} & 59.6 {(-0.7)} & \textbf{44.0 {(+7.3)}} \\
\texttt{Xmath} & 73.3 {(+3.3)} & 47.2 {(+1.1)} & 68.8 {(-0.5)} & 61.0 {(+1.7)} & 40.9 {(-0.3)} \\
\texttt{Xcode} & 73.1 {(+3.0)} & 47.2 {(+1.1)} & 69.3 {(+0.3)} & 60.0 {(+0.0)} & 43.1 {(+5.1)} \\
\midrule
\textbf{Model} & \textbf{MathVista} & \textbf{HellaSwag} & \textbf{HallBench} & \textbf{POPE} & \textbf{MTVQA} \\
\midrule
\texttt{base} & 45.8 {(+0.0)} & 48.9 {(+0.0)} & 53.8 {(+0.0)} & 86.7 {(+0.0)} & \textbf{23.9 {(+0.0)}} \\
\texttt{Xvqa} & 39.9 {(-12.9)} & 51.7 {(+5.6)} & 51.4 {(-4.5)} & 87.4 {(+0.8)} & 22.5 {(-6.0)} \\
\texttt{Xmlingual} & \textbf{48.9 {(+6.8)}} & 48.4 {(-1.2)} & 53.5 {(-0.6)} & 81.5 {(-6.0)} & 20.1 {(-16.1)} \\
\texttt{Xrealworld} & 46.9 {(+2.4)} & \textbf{53.4 {(+9.1)}} & 54.3 {(+0.8)} & 84.0 {(-3.1)} & 23.0 {(-4.0)} \\
\texttt{Xtext} & 47.9 {(+4.6)} & 45.4 {(-7.3)} & 53.7 {(-0.2)} & 87.6 {(+1.0)} & 23.5 {(-1.8)} \\
\texttt{Xocr} & 48.0 {(+4.8)} & 52.8 {(+8.0)} & 55.1 {(+2.3)} & 86.0 {(-0.8)} & 23.1 {(-3.7)} \\
\texttt{Xcharts} & 40.8 {(-10.9)} & 50.8 {(+3.9)} & 52.4 {(-2.7)} & \textbf{88.8 {(+2.4)}} & 23.9 {(-0.1)} \\
\texttt{Xdocs} & 47.3 {(+3.3)} & 50.9 {(+4.0)} & 52.6 {(-2.3)} & 86.1 {(-0.7)} & 23.3 {(-2.6)} \\
\texttt{Xmath} & 43.7 {(-4.6)} & 48.9 {(+0.0)} & \textbf{55.4 {(+2.9)}} & 86.2 {(-0.6)} & 23.2 {(-3.0)} \\
\texttt{Xcode} & 46.9 {(+2.4)} & 43.8 {(-10.6)} & 53.4 {(-0.8)} & 86.0 {(-0.8)} & 23.2 {(-3.0)} \\
\bottomrule
\end{tabular}%
}
\raggedright\small
~~~
\newline
Some evaluation tasks from Section~\ref{sec:evaluation} are omitted due to insignificant variance. Relative differences to the baseline (\%) are shown in parentheses. \textbf{Bold} marks the best score per column. All scores represent accuracy (\%) except OCRBench (0--1000 scale).
\label{tab:ablation-results}
\end{table}

Table~\ref{tab:ablation-results} summarizes ablation results; observations are correlational, based on 10\% of full training, and should be treated as hypotheses rather than established mechanisms.

Removing VQA data (\texttt{Xvqa}) causes dramatic but highly task-specific degradation on VQA benchmarks with minimal impact on other tasks. Similarly, \texttt{Xocr} harms OCRBench while improving all other tasks. This confirms that task-specific data (VQA, OCR) primarily benefits its own tasks without broad transfer. Domain removals show predictable patterns for \texttt{Xcharts} (ChartQA $-39.4\%$) and \texttt{Xmath} (MathVista $-4.6\%$). More surprisingly, removing real-world (\texttt{Xrealworld}) or document data (\texttt{Xdocs}) improves several unrelated benchmarks (HellaSwag $+9.1\%$, MMMU $+7.3\%$), while code removal (\texttt{Xcode}) degrades HellaSwag ($-10.6\%$). Removing text-only data (\texttt{Xtext}) consistently improves multimodal benchmarks but hurts HellaSwag, the only unimodal task. Removing multilingual data (\texttt{Xmlingual}) improves most tasks but, as expected, drops MTVQA by $-16.1\%$.

These results support three conclusions for practitioners under fixed compute budgets. First, task-specific data (VQA, OCR, charts, math) is necessary for its capabilities but does not transfer broadly—consistent with MM1~\citep{mm1} and Cambrian-1~\citep{cambrian1}. Second, real-world and document data exhibit unexpected interference: their removal improves several unrelated benchmarks, suggesting noise or gradient conflict at this scale. Third, text-only and multilingual data impose a cost on multimodal performance but are essential for their respective capabilities; practitioners who do not require them should omit them.


\section{Conclusion}
\label{sec:conclusion}

We presented \jvlm{}, a 2.4B vision-language model that achieves state-of-the-art multilingual VQA and leading English VQA results among open 2B-scale VLMs.

The current approach has limitations. Multi-tile processing introduces computational overhead that scales with image resolution, and tiling can fragment global spatial context, potentially impairing performance on tasks requiring holistic scene understanding such as object counting or precise spatial reasoning across tile boundaries. While the global thumbnail partially mitigates this, native-resolution approaches \citep{navit} may be better suited for such tasks. We have not emphasized safety-critical training or alignment, and multi-image reasoning remains weak due to limited training data in this regime. Future work could explore more efficient resolution handling, targeted improvements for counting and spatial tasks, and investigate whether our multilingual training recipe transfers to larger model scales.

Our data mixture analysis offers diagnostic observations for resource-constrained training; as ablations ran at 10\% of full training, findings should be interpreted as preliminary evidence motivating future systematic investigation.


\newpage

\bibliography{jvlm_datafm_iclr2026}
\bibliographystyle{jvlm_datafm_iclr2026}


\newpage
\appendix
\section{Appendix}
\label{appendix}

\subsection{Pseudocode for Creating Overlapping Tiles}
\label{appendix:cropping-pseudocode}

\begin{algorithm}[H]
\caption{\textsc{GetAllTilesOverlapAndResize}}
\KwIn{%
  Image $I$ of size $(h, w)$; \newline
  Base input size $\mathbf{b} = (b_h, b_w)$ ($(378, 378)$) \newline
  Patch size $p$ ($14$); \newline
  Maximum number of tiles $M$ ($12$ by default, configurable) \newline
  Overlap margins $(m_L, m_R)$ in patches ($(4, 4)$)
}
\KwOut{
  List of tiles $\mathcal{C}$ (thumbnail + grid tiles) \newline
  Tiling $(t_h, t_w)$ = (number of rows, number of columns)
}

\textbf{1. Compute overlap-related sizes}\\
$m_{\text{tot}} \gets p \cdot (m_L + m_R)$ \tcp{Total overlap margin in pixels}
$s_{\text{win}} \gets \bigl(\lfloor b_h / p \rfloor - (m_L + m_R)\bigr) \cdot p$ \tcp{Tile stride in pixels}

\textbf{2. Select tiling on the margin-reduced image}\\
$(t_h, t_w) \gets \textsc{SelectTilingWithMinimalScaleChange}\bigl(
      h - m_{\text{tot}},\,
      w - m_{\text{tot}},\,
      s_{\text{win}},\,
      M \bigr)$\;

\textbf{3. Resize image to exactly fit the chosen tiling + margins}\;
$H' \gets t_h \cdot s_{\text{win}} + m_{\text{tot}}$\;
$W' \gets t_w \cdot s_{\text{win}} + m_{\text{tot}}$\;
$I_{\text{grid}} \gets \textsc{Resize}(I,\; [H', W'])$\;

\textbf{4. Extract overlapping tiles}\\
$\mathcal{G} \gets \textsc{ExtractTiles}\bigl(
      I_{\text{grid}},\,
      (t_h, t_w),\,
      s_{\text{win}},\,
      b_h\bigr)$ \tcp{$b_h$ is the tile height, equal to $b_w$ here}

\textbf{5. Build thumbnail and final tile list}\\
$T \gets \textsc{Resize}(I,\; [b_h, b_w])$ \tcp{Global thumbnail}
$\mathcal{C} \gets [T] \mathbin{+\!\!+} \mathcal{G}$ \tcp{Concatenate thumbnail and tiles}
\Return $(\mathcal{C}, (t_h, t_w))$\;

\end{algorithm}

\newpage

\subsection{Training Set Examples}
\label{appendix:examples}

\begin{figure}[H]
\begin{sftpanel}{Captioning \& Instruction}
  \textbf{Dataset:} \ds{VisualWebInstruct}%
  \hfill {\footnotesize\cite{visualwebinstruct}}\\[0.75em]
  \begin{minipage}{0.42\textwidth}
    \centering
    \includegraphics[width=\linewidth]{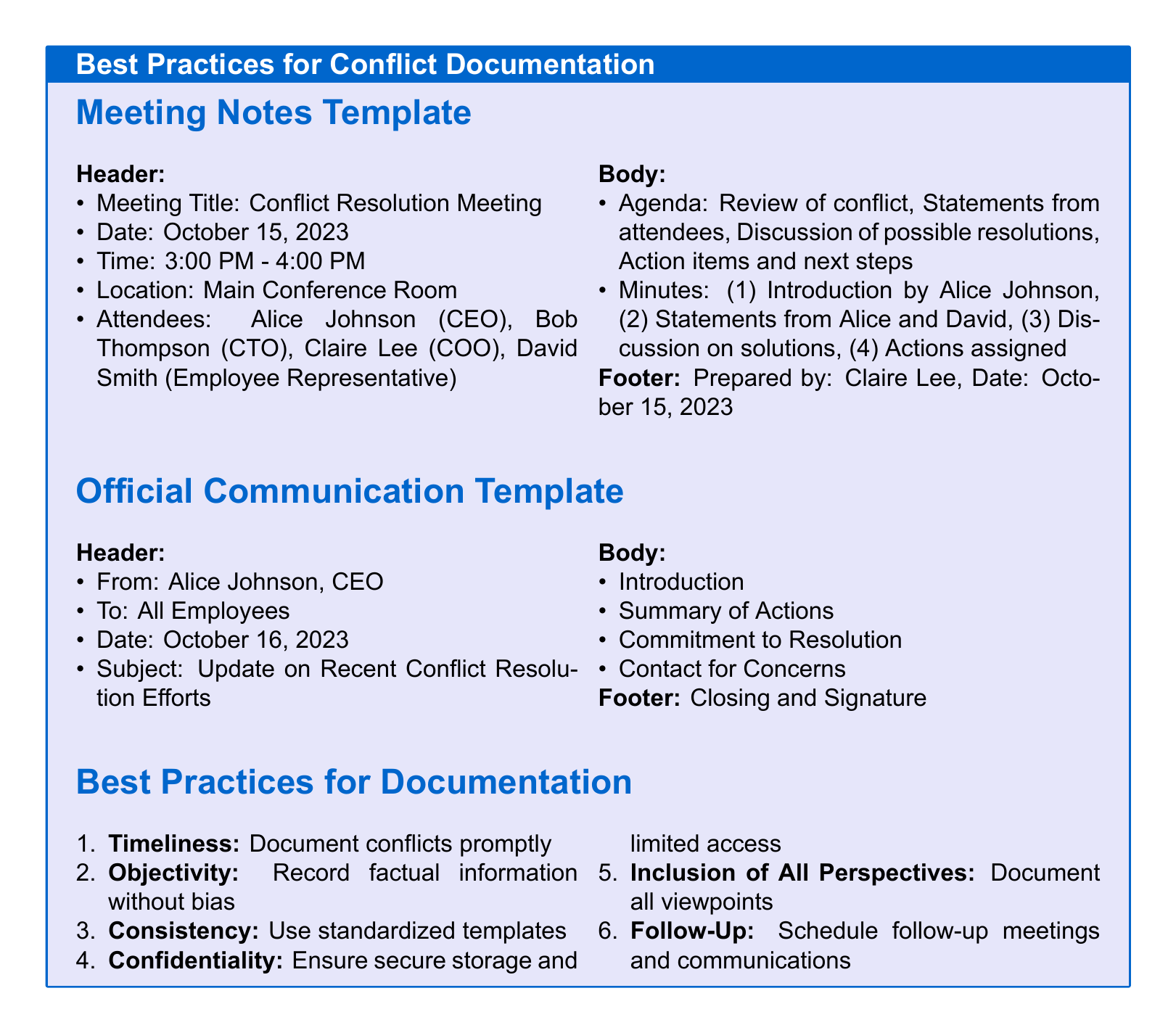}
  \end{minipage}\hfill
  \begin{minipage}{0.54\textwidth}
    {\bfseries Question}\\[0.25em]
    what is the meeting title?\\[0.75em]
    {\bfseries Answer}\\[0.25em]
    Conflict Resolution Meeting
  \end{minipage}
\end{sftpanel}
\caption{Answer questions given web documents.}
\end{figure}

\begin{figure}[H]
\begin{sftpanel}{Charts \& Tables}
  \textbf{Dataset:} \ds{TAT-QA}%
  \hfill {\footnotesize\cite{tatqa}}\\[0.75em]
  \begin{minipage}{0.42\textwidth}
    \centering
    \includegraphics[width=\linewidth]{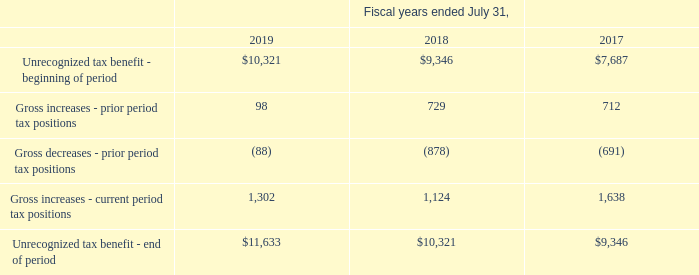}
  \end{minipage}\hfill
  \begin{minipage}{0.54\textwidth}
    {\bfseries Question}\\[0.25em]
    Unrecognized Tax Benefits Activity related to unrecognized tax benefits is as follows (in thousands): ... As of July 31, 2019, the Company has no income tax audits in progress in the U.S. or foreign jurisdictions. What was the increase in unrecognized tax benefits in 2019?\\[0.75em]
    {\bfseries Answer}\\[0.25em]
    \$1.3 million.
  \end{minipage}
\end{sftpanel}
\caption{Financial table requiring numerical reasoning over text.}
\end{figure}

\begin{figure}[H]
\begin{sftpanel}{Document Understanding \& Infographics}
  \textbf{Dataset:} \ds{DocVQA}%
  \hfill {\footnotesize\cite{docvqa}}\\[0.75em]
  \begin{minipage}{0.42\textwidth}
    \centering
    \includegraphics[width=\linewidth]{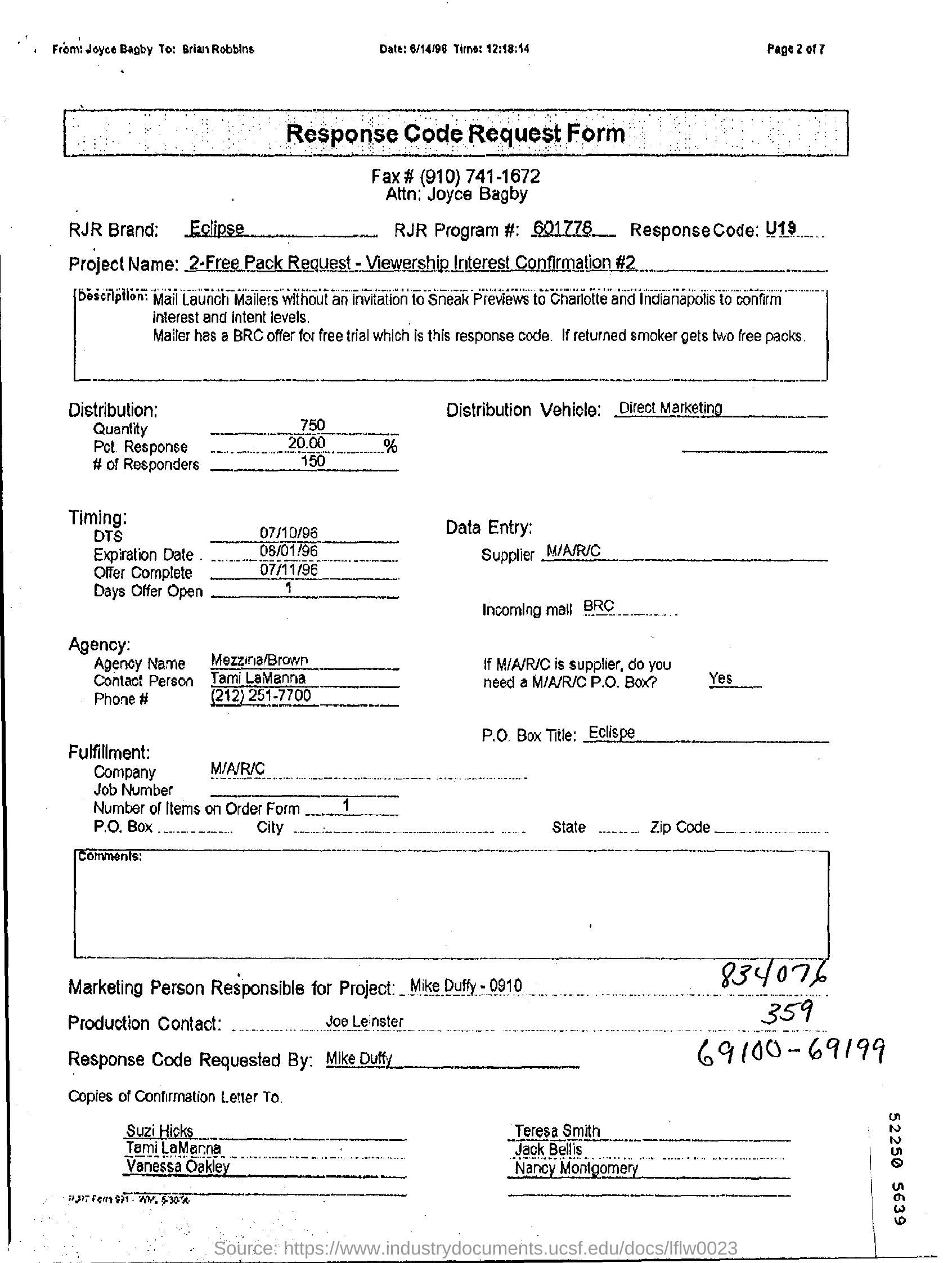}
  \end{minipage}\hfill
  \begin{minipage}{0.54\textwidth}
    {\bfseries Question}\\[0.25em]
    what is the response code ?\\[0.75em]
    {\bfseries Answer}\\[0.25em]
    U19
  \end{minipage}
\end{sftpanel}
\caption{Document image with question about textual fields.}
\end{figure}

\begin{figure}[H]
\begin{sftpanel}{OCR QA (text-centric VQA)}
  \textbf{Dataset:} \ds{TextVQA}%
  \hfill {\footnotesize\cite{textvqa}}\\[0.75em]
  \begin{minipage}{0.42\textwidth}
    \centering
    \includegraphics[width=\linewidth]{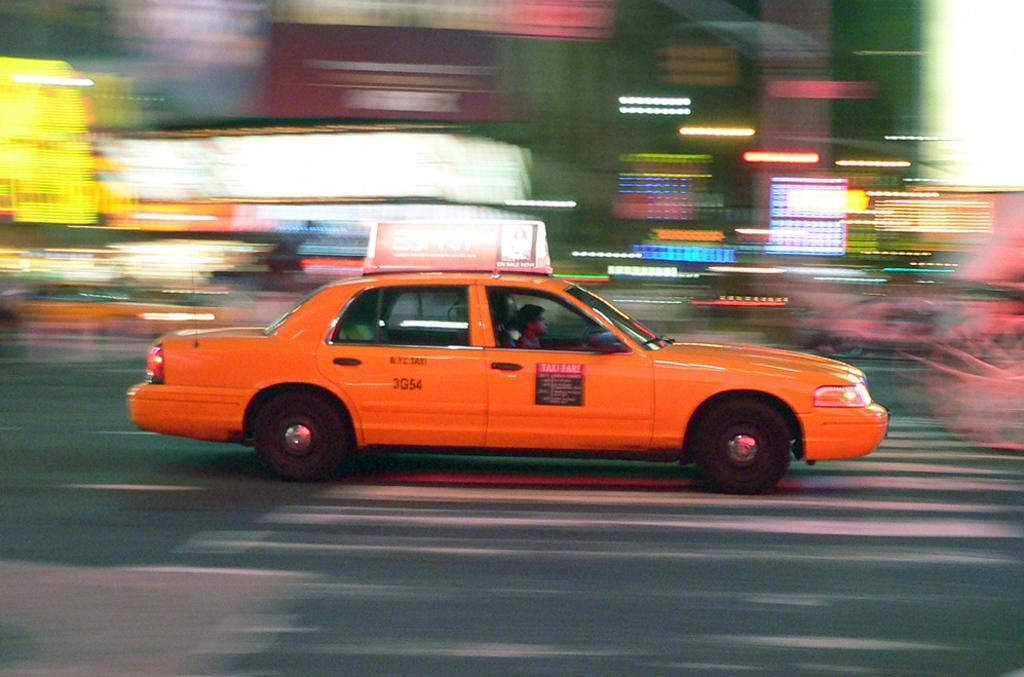}
  \end{minipage}\hfill
  \begin{minipage}{0.54\textwidth}
    {\bfseries Question}\\[0.25em]
    what number is the cab\\[0.75em]
    {\bfseries Answer}\\[0.25em]
    3G54
  \end{minipage}
\end{sftpanel}
\caption{Photo with textual question needing OCR reading.}
\end{figure}

\begin{figure}[H]
\begin{sftpanel}{General VQA}
  \textbf{Dataset:} \ds{VQAv2}%
  \hfill {\footnotesize\cite{vqav2}}\\[0.75em]
  \begin{minipage}{0.42\textwidth}
    \centering
    \includegraphics[width=\linewidth]{selected_samples/vqav2.png}
  \end{minipage}\hfill
  \begin{minipage}{0.54\textwidth}
    {\bfseries Question}\\[0.25em]
    Where is he looking?\\[0.75em]
    {\bfseries Answer}\\[0.25em]
    down
  \end{minipage}
\end{sftpanel}
\caption{General visual question answering on natural images.}
\end{figure}

\begin{figure}[H]
\begin{sftpanel}{Grounding, Spatial \& Counting}
  \textbf{Dataset:} \ds{TallyQA}%
  \hfill {\footnotesize\cite{tallyqa}}\\[0.75em]
  \begin{minipage}{0.42\textwidth}
    \centering
    \includegraphics[width=\linewidth]{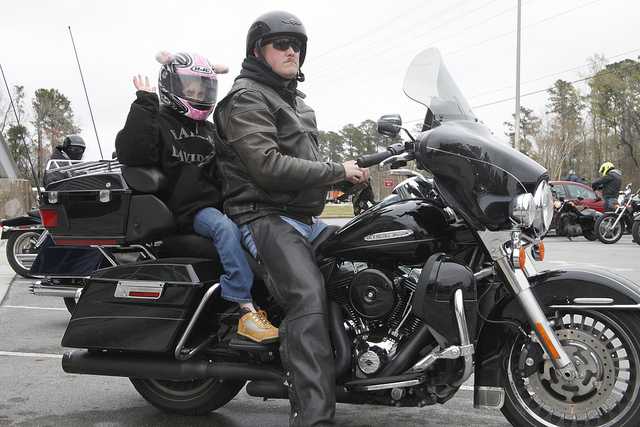}
  \end{minipage}\hfill
  \begin{minipage}{0.54\textwidth}
    {\bfseries Question}\\[0.25em]
    How many more people can ride on the vehicle?\\[0.75em]
    {\bfseries Answer}\\[0.25em]
    0
  \end{minipage}
\end{sftpanel}
\caption{Scene requiring counting and spatial reasoning accuracy.}
\end{figure}

\begin{figure}[H]
\begin{sftpanel}{Math \& Geometry (vision)}
  \textbf{Dataset:} \ds{CLEVR}%
  \hfill {\footnotesize\cite{clevr}}\\[0.75em]
  \begin{minipage}{0.42\textwidth}
    \centering
    \includegraphics[width=\linewidth]{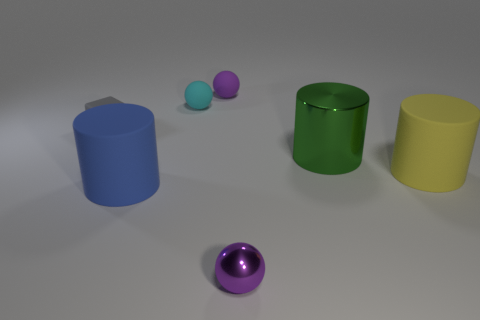}
  \end{minipage}\hfill
  \begin{minipage}{0.54\textwidth}
    {\bfseries Question}\\[0.25em]
    There is a large shiny object; does it have the same shape as the object right of the large metallic thing? Provide a short and direct response.\\[0.75em]
    {\bfseries Answer}\\[0.25em]
    Yes.
  \end{minipage}
\end{sftpanel}
\caption{Synthetic shapes testing compositional spatial reasoning.}
\end{figure}

\begin{figure}[H]
\begin{sftpanel}{Screens, Web \& GUI}
  \textbf{Dataset:} \ds{ScreenQA}%
  \hfill {\footnotesize\cite{screenqa}}\\[0.75em]
  \begin{minipage}{0.42\textwidth}
    \centering
    \includegraphics[width=\linewidth]{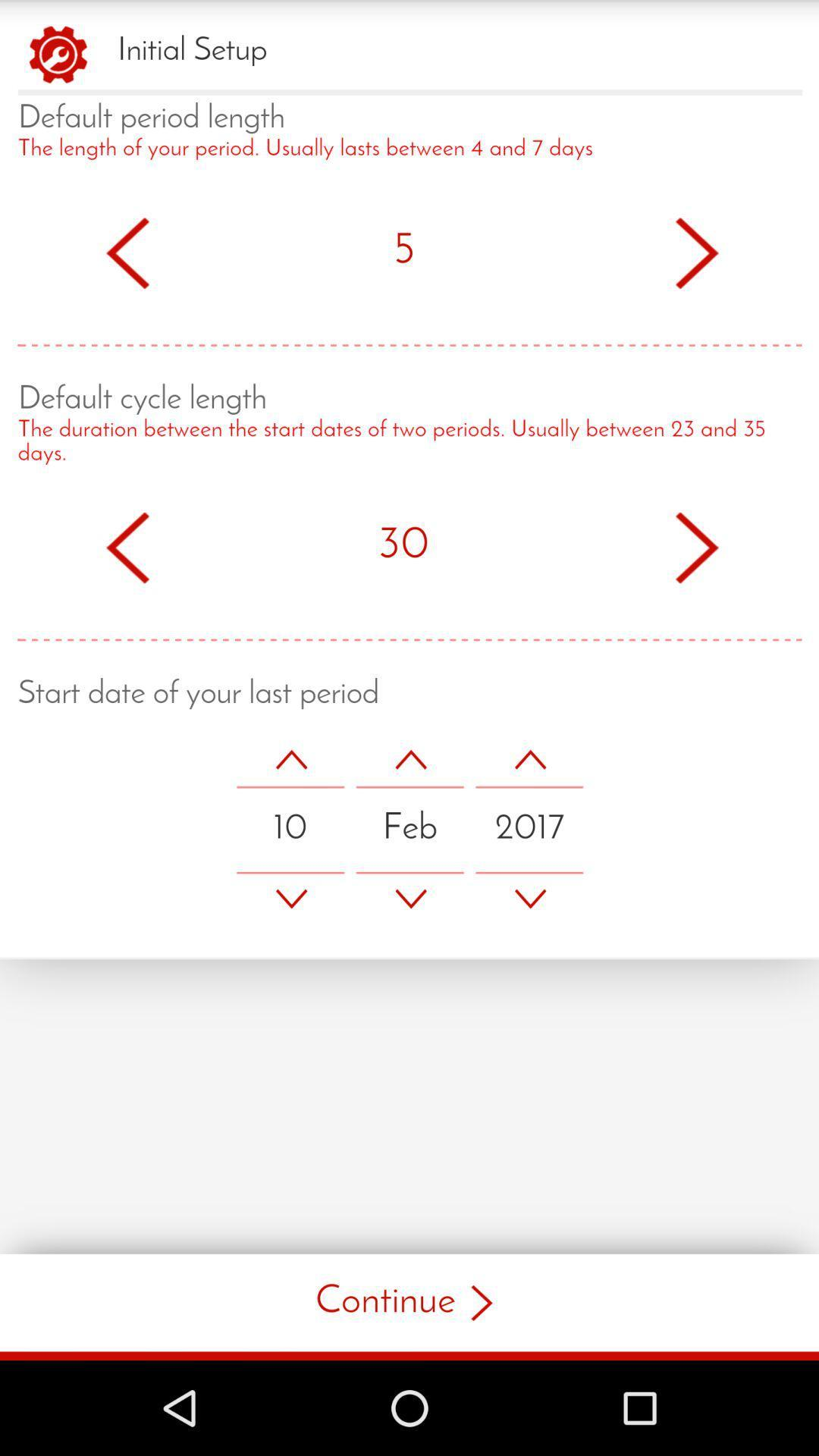}
  \end{minipage}\hfill
  \begin{minipage}{0.54\textwidth}
    {\bfseries Question}\\[0.25em]
    What is the default period length?\\[0.75em]
    {\bfseries Answer}\\[0.25em]
    5 days
  \end{minipage}
\end{sftpanel}
\caption{User interface screenshot with structured textual elements.}
\end{figure}

\begin{figure}[H]
\begin{sftpanel}{Medical (vision)}
  \textbf{Dataset:} \ds{PathVQA}%
  \hfill {\footnotesize\cite{pathvqa}}\\[0.75em]
  \begin{minipage}{0.42\textwidth}
    \centering
    \includegraphics[width=\linewidth]{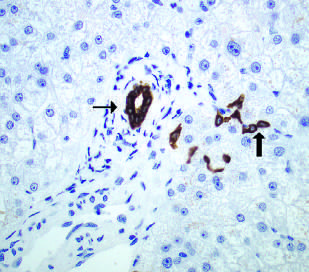}
  \end{minipage}\hfill
  \begin{minipage}{0.54\textwidth}
    {\bfseries Question}\\[0.25em]
    where are liver stem cells (oval cells) located?\\[0.75em]
    {\bfseries Answer}\\[0.25em]
    in the canals of hering
  \end{minipage}
\end{sftpanel}
\caption{Microscopic pathology image for medical VQA.}
\end{figure}

\begin{figure}[H]
\begin{sftpanel}{Text-only (instruction / code / math / translation)}
  \textbf{Dataset:} \ds{aya\_dataset}%
  \hfill {\footnotesize\cite{aya}}\\[0.75em]
  {\bfseries Question}\\[0.25em]
  Quels président des États-Unis ne s’est jamais marié?
  \\[0.75em]
  {\bfseries Answer}\\[0.25em]
  James Buchanan est le seul président qui ne s'est jamais marié.
\end{sftpanel}
\caption{Text-only tasks covering multiple languages.}
\end{figure}

\newpage

\subsection{Instruction Fine-Tuning Stage Dataset Composition}
\label{appendix:composition-finetuning}

\begin{table}[htpb]
\centering
\caption{Instruction Fine-Tuning dataset composition}
\begin{tabular}{lr}
\toprule
\textbf{Category} & \textbf{Percentage (\%)} \\
\midrule
\multicolumn{2}{l}{\textbf{Tasks}} \\
VQA & 29.08 \\
OCR & 17.92 \\
QA & 15.88 \\
Open-ended VQA & 8.57 \\
Reasoning QA & 8.49 \\
Instruction Following & 8.26 \\
Pointing VQA & 4.05 \\
Captions & 2.36 \\
Long Captions & 2.09 \\
Counting VQA & 1.08 \\
Action/Instruction Following & 0.82 \\
Navigation & 0.64 \\
Classification & 0.28 \\
Misc & 0.21 \\
Sentiment Analysis & 0.20 \\
Multi Image Reasoning & 0.07 \\
\midrule
\multicolumn{2}{l}{\textbf{Domains}} \\
Real World & 24.53 \\
Math & 12.46 \\
Charts \& Diagrams & 12.14 \\
Web & 12.00 \\
Documents & 7.20 \\
Code & 5.93 \\
Misc & 5.15 \\
Screens \& GUI & 3.97 \\
Geometry & 3.91 \\
Tables & 3.15 \\
Handwritten & 2.02 \\
Synthetic Scenes & 1.16 \\
Landmarks & 1.11 \\
Infographics & 1.02 \\
Medical & 0.92 \\
Captcha & 0.68 \\
Academic & 0.64 \\
Science & 0.64 \\
Maps & 0.49 \\
Rendered Text & 0.29 \\
Books & 0.26 \\
Receipts & 0.18 \\
Comics & 0.13 \\
\midrule
\multicolumn{2}{l}{\textbf{Modalities}} \\
Image+Text & 82.21 \\
Text & 17.79 \\
\midrule
\multicolumn{2}{l}{\textbf{Languages}} \\
English & 85.17 \\
Multilingual & 14.83 \\
\bottomrule
\end{tabular}
\label{tab:composition-finetuning}
\end{table}

\newpage

\begin{table}[t!]
\centering
\caption{Task type reference table}
\resizebox{\textwidth}{!}{%
\begin{tabular}{lp{9.5cm}}
\toprule
\textbf{Task Type} & \textbf{Description} \\
\midrule
\textbf{VQA} & Visual question answering - questions about the image \\
\textbf{OCR} & Optical character recognition \\
\textbf{QA} & Question answering where there is no image or the image does not provide the knowledge needed to answer the question \\
\textbf{Open-ended VQA} & Open-ended questions e.g. "Describe the scene" or "What is happening in the picture?" \\
\textbf{Reasoning QA} & Question answering with a specific request to return the reasoning \\
\textbf{Instruction Following} & Following instructions given in the text \\
\textbf{Pointing VQA} & Questions requiring localization or pointing to specific regions \\
\textbf{Captions} & Short image descriptions (1-2 sentences) \\
\textbf{Long Captions} & Detailed image descriptions (multiple sentences, comprehensive) \\
\textbf{Counting VQA} & Questions requiring counting objects in images \\
\textbf{Action/Instruction Following} & Understanding and following action-based instructions \\
\textbf{Navigation} & Understanding spatial relationships and navigation \\
\textbf{Classification} & Categorizing images into predefined classes \\
\textbf{Sentiment Analysis} & Analyzing sentiment from combined image and text \\
\textbf{Multi Image Reasoning} & Reasoning across multiple images \\
\bottomrule
\end{tabular}%
}
\label{tab:task-reference}
\end{table}

\begin{table}[t!]
\centering
\caption{Domain reference table}
\resizebox{\textwidth}{!}{%
\begin{tabular}{ll}
\toprule
\textbf{Domain} & \textbf{Description} \\
\midrule
\textbf{Real World} & Natural photos of real-world scenes and objects (landscapes, people, animals) \\
\textbf{Math} & Math problems, equations \\
\textbf{Charts \& Diagrams} & Data visualizations, flowcharts, graphs, plots \\
\textbf{Web} & Websites, memes, AI-generated images and generic web content \\
\textbf{Documents} & General document images (reports, forms) \\
\textbf{Code} & Code snippets, technical documentation \\
\textbf{Misc} & Miscellaneous content, various domains not covered elsewhere \\
\textbf{Screens \& GUI} & Computer screens, mobile apps, software interfaces \\
\textbf{Geometry} & Geometric problems, trigonometry, proofs, mathematical drawings \\
\textbf{Tables} & Structured tabular data \\
\textbf{Handwritten} & Handwritten text \\
\textbf{Synthetic Scenes} & Computer-generated depictions of real-world scenes \\
\textbf{Landmarks} & Famous buildings, monuments, tourist attractions \\
\textbf{Infographics} & Graphics combining text and visuals (event posters, leaflets) \\
\textbf{Medical} & Healthcare images, anatomy, medical scans, clinical documents \\
\textbf{Captcha} & CAPTCHA verification images \\
\textbf{Academic} & Research articles and content from various academic domains \\
\textbf{Science} & Scientific images, experiments, lab setups, diagrams, educational content \\
\textbf{Maps} & Geographic maps, navigation, spatial information \\
\textbf{Rendered Text} & Computer-generated text overlays and typography \\
\textbf{Books} & Book pages, literature \\
\textbf{Receipts} & Purchase receipts, invoices, transaction records \\
\textbf{Comics} & Comic books, manga, graphic novels \\
\bottomrule
\end{tabular}%
}
\label{tab:domain-reference}
\end{table}

\end{document}